\documentclass[letterpaper, 10 pt, conference]{ieeeconf}

\IEEEoverridecommandlockouts
\usepackage{graphicx}
\usepackage{epsfig}
\usepackage{mathptmx}
\usepackage{times}
\usepackage{amsmath}
\usepackage{amssymb}
\usepackage[T1]{fontenc}
\usepackage[utf8]{inputenc}
\usepackage{booktabs}
\usepackage{tabularx}
\usepackage{multirow}
\usepackage{array}
\usepackage{url,comment,color}
\usepackage{cite}
\usepackage{hyperref}

\title{\LARGE \bf
Multi-Field Hybrid Retrieval-Augmented Generation for Maritime Accident Root Cause Analysis
}

\author{Seongjin Kim$^{a}$ and Sungil Kim$^{a,\dagger}$%
  \thanks{$^{a}$Department of Industrial Engineering,
    Ulsan National Institute of Science and Technology (UNIST), Ulsan 44919, Republic of Korea.}%
  \thanks{$^\dagger$Corresponding author:
    \href{mailto:sungil.kim@unist.ac.kr}{sungil.kim@unist.ac.kr}}%
}

\begin{document}
\maketitle
\thispagestyle{empty}
\pagestyle{empty}

\begin{abstract}
Maritime accident adjudication reports contain critical tribunal findings for root cause analysis (RCA), yet retrieving relevant precedents and drafting consistent reports from decades of records remains labor-intensive. This paper proposes a multi-field hybrid retrieval-augmented generation (RAG) framework for automated maritime RCA, utilizing a comprehensive dataset of 13,329 Korea Maritime Safety Tribunal (KMST) reports (1971--2025). We transform raw adjudications into a structured knowledge base of ``incident cards'', indexing three distinct fields—\textit{Summary}, \textit{Causes}, and \textit{Disposition}—alongside a hierarchical L1/L2 cause taxonomy. Our retrieval strategy employs a field-aware hybrid approach, fusing sparse and dense rankings via Reciprocal Rank Fusion (RRF). Given the lack of large-scale expert relevance labels, we evaluate retrieval performance using ceiling-normalized recall and nDCG based on a metadata-derived proxy relevance score. Experimental results demonstrate that our proposed retrieval significantly outperforms baseline methods, improving NormRecall@100 from 0.18 to 0.55. Furthermore, grounding the generator on the retrieved precedents enhances RCA generation quality over an LLM-only baseline, increasing the LLM-as-a-judge score from 3.34 to 3.72. These findings suggest that field-aware RAG can substantially streamline maritime safety investigation workflows by enabling faster precedent search and more consistent, evidence-based RCA drafting.
\end{abstract}

\section{INTRODUCTION}

Maritime accidents can lead to severe safety consequences and operational disruptions, including delays and capacity losses in maritime transport and port logistics~\cite{ART002614209,oh2023grid,oh2025comparative}.
To prevent recurrence, systematic Root Cause Analysis (RCA) is essential for robust safety management \cite{reason1990human,fan2020review}. In practice, RCA often relies on analogical reasoning over past adjudications: investigators search for historically similar cases, examine tribunal cause findings, and justify conclusions with precedent-based evidence.

\begin{figure}[t]
  \centering
  \includegraphics[width=0.83\columnwidth]{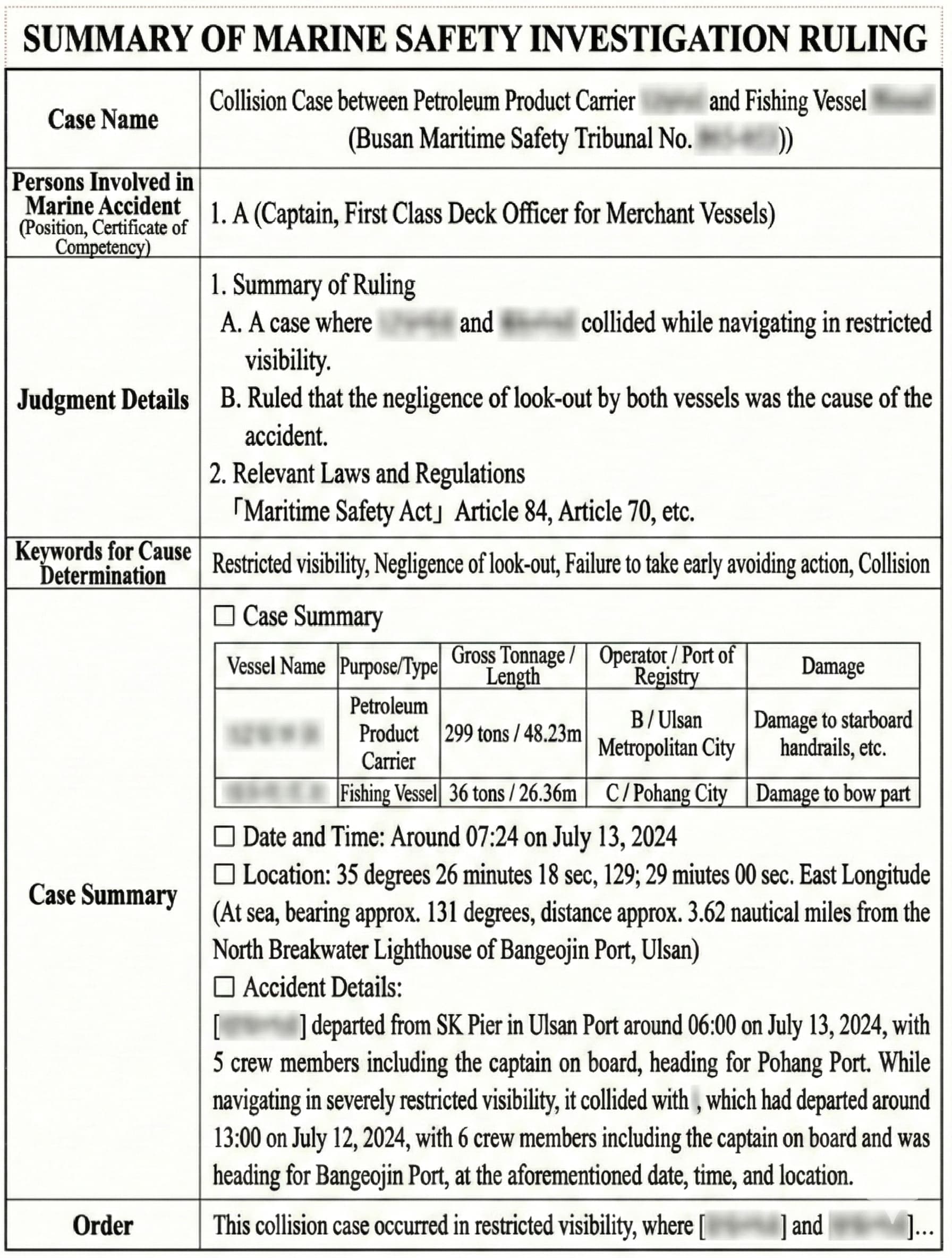}
  \caption{Example KMST adjudication report (translated into English from the original Korean, identifiers blurred)}
  \label{fig:raw_example}
  \vspace{-0.1 in}
\end{figure}

Despite its importance, performing precedent-based search over formal adjudication reports (e.g., Fig.~\ref{fig:raw_example}) remains a daunting task. These records span decades and are characterized by a dense mixture of domain-specific legal and technical terminology. Furthermore, the information is fragmented across multiple sections, such as accident narratives, causal reasoning, and administrative dispositions. Consequently, analysts spend substantial time manually synthesizing these disparate parts, which hinders timely organizational learning and consistent safety intervention.

Recent advances in Large Language Models (LLMs) offer a potential solution for automating such narrative synthesis. However, in specialized legal--technical domains, LLMs are prone to \textit{hallucinations}—generating plausible but factually incorrect regulations or causal mechanisms. While Retrieval-Augmented Generation (RAG) mitigates this by conditioning outputs on external evidence \cite{lewis2020rag,gao2023rag_survey}, its application to maritime RCA faces two distinct challenges. First, adjudication reports are \textbf{multi-field documents} with diverse semantics (\textit{Summary}, \textit{Causes}, \textit{Disposition}); treating them as monolithic text blocks dilutes the field-specific signals necessary for precise retrieval~\cite{soh2018application}. Second, \textbf{evaluation at scale} is notoriously difficult, as expert relevance judgments are expensive to obtain, yet retrieval quality is the primary determinant of downstream reliability.

In this work, we formulate maritime RCA assistance as \textbf{case-informed structured generation}. Given a new accident scenario, our system retrieves historically similar adjudications and generates a structured RCA output grounded in these precedents. We specifically focus on leveraging multi-field document structures to enhance retrieval precision and introduce a reproducible evaluation proxy for large-scale benchmarking.

The main contributions of this study are as follows:
\begin{itemize}
    \item \textbf{Curating a Structured Maritime Knowledge Base:} We construct a large-scale, structured knowledge base of 13,329 KMST adjudication cases (1971--2025), featuring retrievable document fields and hierarchical L1/L2 cause tags to facilitate evidence-grounded analysis.
    
    \item \textbf{Proposing an Integrated Retrieval and Evaluation Framework:} We develop a multi-field hybrid retrieval strategy that optimizes case matching and introduce a reproducible, metadata-driven evaluation protocol that demonstrates significantly improved structured RCA generation over LLM-only baselines.
\end{itemize}

\section{RELATED WORK}

\subsection{RCA and Precedent-Grounded Investigation}
RCA in safety-critical systems seeks to explain how multiple contributing factors combine to produce adverse events \cite{reason1990human,fan2020review}. In maritime safety, tribunal-style adjudication reports encode expert RCA judgments—comprising causal reasoning and administrative dispositions—that investigators consult as precedents. This motivates treating adjudication records as an evidence corpus for retrieval-based analysis, where a system must not only \emph{find} relevant precedents but also \emph{synthesize} explanations with traceable support.
\subsection{Specialized RAG and Hybrid Retrieval}
RAG reduces hallucination in knowledge-intensive tasks by grounding LLM outputs in external evidence \cite{lewis2020rag,gao2023rag_survey}. For semi-structured technical reports, retrieval relevance is multifaceted: it involves descriptive similarity in narratives, causal similarity in reasoning, and outcome similarity in dispositions. While sparse lexical methods (e.g., BM25) excel at matching rare technical terms \cite{robertson2009bm25}, dense retrieval captures semantic nuances using embeddings like \texttt{BAAI/bge-m3} \cite{chen2024bgem3}. Hybrid retrieval, fused via Reciprocal Rank Fusion (RRF), combines these strengths \cite{cormack2009rrf,Kim2026hybrid}. Our work extends this by utilizing \textit{field-aware} indexing \cite{macavaney2020structured}, preventing the dilution of distinct causal signals that occurs in monolithic text blocks~\cite{kim2023contextual}. 

\subsection{Weak Supervision for Domain Taxonomies}
In specialized domains, large-scale expert annotation is often infeasible, necessitating rule-based information extraction and weak supervision to generate structured signals \cite{ratner2017snorkel}. Taxonomy-aligned metadata enables standardized outputs and corpus-level analysis. Following these principles, we employ a rule-based mapping to assign KMST L1/L2 cause tags, providing a consistent label space for both the structured knowledge base and the generator $\mathcal{G}$.
\subsection{Maritime Accident Text Mining}
Prior research on maritime reports has addressed various analytical tasks, including accident type prediction \cite{park2020developing}, statistical cause analysis \cite{choi2021marine,lee2023quantifying}, causal relation extraction \cite{yan2023causation,moon2023sequence}, semantic clustering \cite{yoon2023sbert}, and knowledge graph construction \cite{MAKG_2024}. While these studies demonstrate the utility of NLP for tribunal records, they are primarily \textit{discriminative} or focused on extracting local patterns from historical data. Such approaches lack the capacity to generate coherent, evidence-grounded explanations for \textit{new} scenarios. In contrast, our framework performs \textit{generative} RCA, synthesizing a causal narrative and standardized tags while explicitly surfacing supporting precedents for traceable investigation.

\section{METHOD}
\subsection{Problem Formulation}
We define the task of maritime RCA assistance as \textit{case-informed structured generation}. Given a query accident scenario $q$—typically a free-text description of a vessel incident—the system retrieves a ranked set of $k$ precedent chunks from a structured knowledge base $\mathcal{D}$:
\begin{equation}
C_k = \text{Retriever}(q, \mathcal{D}) = \{c_1, c_2, \dots, c_k\},
\end{equation}
where each chunk $c_i \in \mathcal{D}$ represents a specific semantic section from a historical adjudication report, such as the \textit{Summary}, \textit{Causes}, or \textit{Disposition} field. For instance, a retrieved chunk $c_1$ may contain the causal reasoning for a past collision similar to $q$.

Conditioned on the input $q$ and the retrieved evidence $C_k$, the generator $\mathcal{G}$ produces a structured RCA output $Y$:
\begin{equation}
Y = \mathcal{G}(q, C_k) = \{n, T\},
\end{equation}
where $n$ is a concise cause narrative and $T$ is a set of taxonomy-aligned cause tags. This formulation ensures that the generation is not only structurally consistent but also grounded in verified historical precedents.

\begin{figure*}[t]
  \centering
  \includegraphics[width=0.7\textwidth]{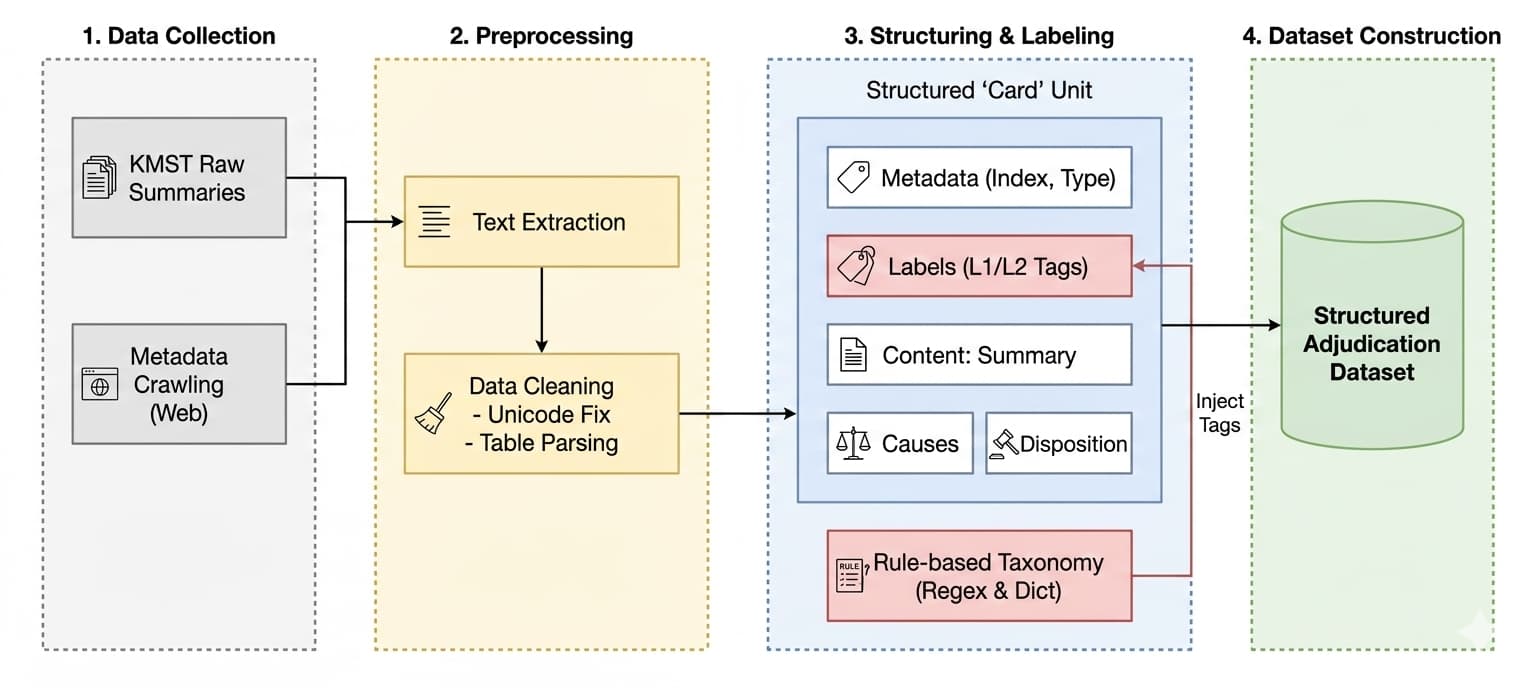}
  \caption{Data construction pipeline: from raw document extraction to structured card normalization and hierarchical L1/L2 tagging.}
  \label{fig:pipeline}
\end{figure*}

\subsection{KMST Adjudication Knowledge Base}
The knowledge base $\mathcal{D}$ is constructed from official adjudication reports published by the Korea Maritime Safety Tribunal (KMST) \cite{kmst_portal}. These records, spanning decades of maritime history, are semi-structured tribunal documents characterized by standardized headings but highly variable phrasing and narrative lengths. As illustrated in Fig.~\ref{fig:pipeline}, we converted each raw record into plain text and applied cleaning and normalization. Each processed record was then standardized into a \textit{card} identified by a unique ID, integrating the textual content with essential metadata such as accident type, keywords, tribunal jurisdiction, and incident year. To preserve semantic specificity, the main text of each card was partitioned into three distinct retrievable fields: \textit{Summary} (incident description), \textit{Causes} (tribunal's causal reasoning), and \textit{Disposition} (administrative outcomes). Each field was stored as an independently retrievable \textit{chunk} while maintaining a reference to its parent case ID for traceability. Across 13,329 standardized cards, this process yielded 37,007 retrievable chunks with an average length of 283 characters (median: 212), providing a dense and structured evidence pool for the subsequent retrieval stage.

\subsection{Cause Taxonomy and Rule-Based Tagging}
To facilitate structured generation, we adopted the KMST cause taxonomy as a standardized label space, which consists of high-level groups (L1) and 24 fine-grained factors (L2). Since raw reports often lack exhaustive ground-truth tags, we employed a rule-based labeling system following the principles of weak supervision \cite{ratner2017snorkel,lee2013dependence}. This system utilizes approximately 2,400 dictionary entries and 800 regex patterns to map the \textit{Keywords} and \textit{Causes} fields of each card to the L1/L2 taxonomy. The labeling process achieved a coverage of 97.6\% (13,012 cases), and the resulting tag distribution aligns with official maritime statistics \cite{kmst_stats_2025}, ensuring the face validity of our structured knowledge base.

\begin{figure*}[t]
\centering
\includegraphics[width=0.85\textwidth]{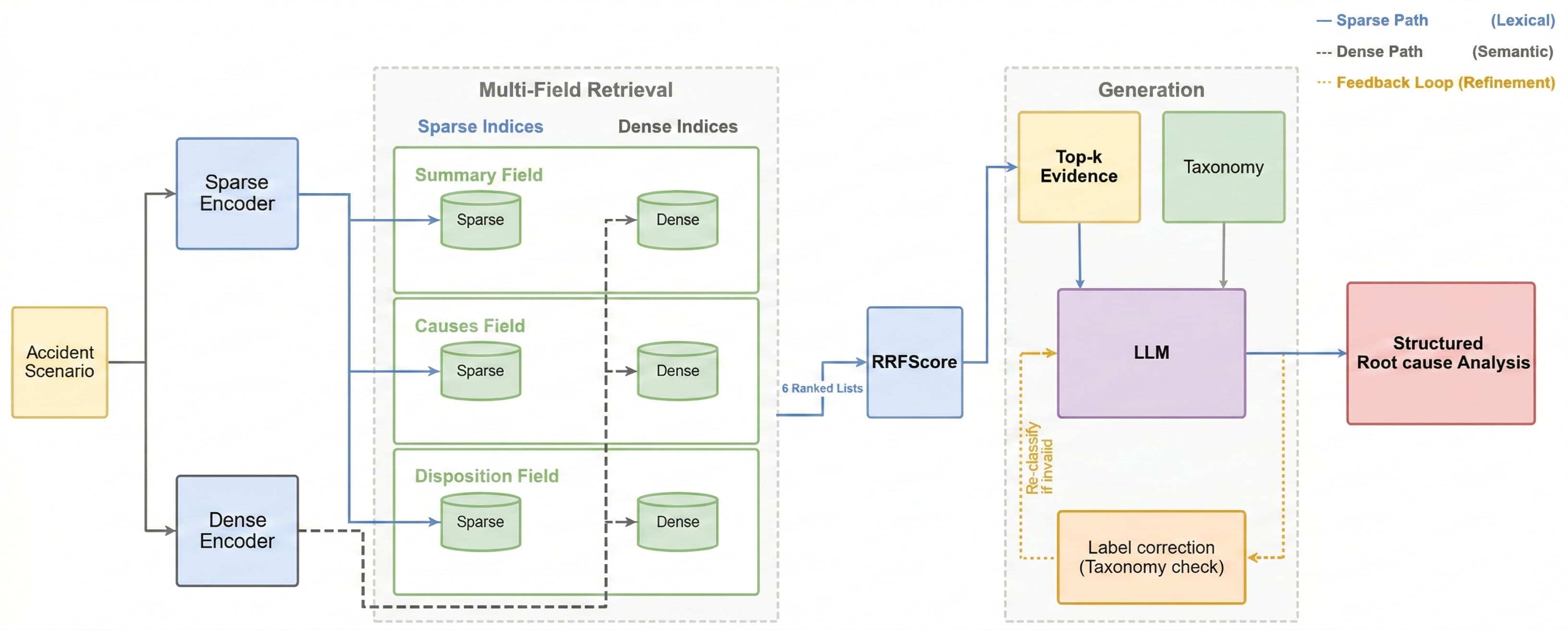}
\caption{Overview of the proposed multi-field hybrid RAG framework.}
\label{fig:framework}
\end{figure*}

\subsection{Multi-Field Hybrid Retrieval}
Given an input scenario $q$, our retriever employs a hybrid strategy designed to capture both lexical overlaps and semantic nuances across different document sections. For each field in $\mathcal{D}$, we implement two complementary retrievers: (1) \textbf{Sparse Retrieval} using BM25 \cite{robertson2009bm25} for exact keyword matching of technical identifiers, and (2) \textbf{Dense Retrieval} using \texttt{bge-m3} \cite{chen2024bgem3} for capturing semantic paraphrases.

To aggregate results across the three fields and two retrieval methods—totaling six independent rankings—we utilize Reciprocal Rank Fusion (RRF) \cite{cormack2009rrf}:
\begin{equation}\label{eq:rrf}
\text{RRFScore}(c) = \sum_{r \in R} \frac{1}{k_{\text{rrf}} + \text{rank}_r(c)},
\end{equation}
where $R$ is the set of rankings and $k_{\text{rrf}}$ is a smoothing constant set to 60. This field-aware fusion ensures that the system recovers relevant evidence even when the query $q$ shares vocabulary with different semantic sections of a historical card.

\subsection{Structured RCA Generation}
The final stage of our framework involves generating the structured output $Y = \{n, T\}$ by conditioning the generator on the retrieved evidence. As illustrated in Fig.~\ref{fig:framework}, the top-$k$ chunks $C_k$ are first sorted in descending order of their RRFScore and then provided as \textit{context} to the generator $\mathcal{G}$ (\texttt{Qwen3-Next-80B-A3B-Instruct}). To maintain traceability and allow the model to distinguish between different types of evidence, each chunk is annotated with its original case ID and field type (\textit{Summary}, \textit{Causes}, or \textit{Disposition}).

We utilize a temperature of 0.0 to minimize output variance and enforce a strict JSON schema via system prompting. The model is specifically tasked with synthesizing the disparate retrieved precedents into a coherent and concise cause narrative $n$, while simultaneously mapping the incident to the appropriate taxonomy-aligned tags $T$. This structured approach ensures that the resulting RCA report is not only human-readable but also consistent with the established KMST cause classification system.

\section{EXPERIMENTS}
\subsection{Setup and Baselines}
We partitioned the 13,329 cards into train, validation, and test sets with a 70\%/15\%/15\% ratio. All retrieval indices were constructed using the training set. Strategy comparisons and hyperparameter tuning (e.g., $k_{\text{rrf}}$) were performed on the validation set, while final performance is reported on the held-out test set.

\textbf{Retrieval Baselines.} To evaluate the efficacy of our multi-field hybrid strategy, we compared it against:
\begin{itemize}
    \item \textbf{Whole-document indexing:} Concatenating the \textit{Summary}, \textit{Causes}, and \textit{Disposition} fields into a single monolithic text block per card.
    \item \textbf{Single-field indexing:} Indexing only the \textit{Summary} or \textit{Causes} fields to assess the information gain from field separation.
\end{itemize}
For each indexing strategy, we evaluated sparse retrieval (BM25), dense retrieval (\texttt{BAAI/bge-m3}), and the proposed hybrid retrieval. Unless otherwise specified, we retrieved $k=100$ chunks for evaluation.

\textbf{Generation Baselines.} We compared our RAG framework against an \textbf{LLM-only baseline}, which utilizes the same generator $\mathcal{G}$ (\texttt{Qwen3-Next-80B-A3B-Instruct}) and prompt structure but without any retrieved context $C_k$. This comparison isolates the impact of precedent-grounding on generation quality. For our proposed system, $\mathcal{G}$ receives the top $k=40$ chunks as context; given our field-aware indexing, this typically provides evidence from 10--13 distinct historical cards.

\textbf{Tag Normalization.} Despite a temperature of 0.0, LLMs may occasionally produce non-standard surface forms for tags instead of exact taxonomy terms. To ensure consistent evaluation of $T$, we applied a lightweight normalization step where $\mathcal{G}$ maps any non-standard output to the closest standardized L2 tag. This was applied identically to both the baseline and the proposed method.

\subsection{Evaluation Protocol}

In the absence of large-scale expert relevance labels for retrieval, we define a proxy relevance metric, $\text{RelScore}$, to evaluate the alignment between a retrieved candidate card $c$ and a query card $q$. This score is calculated as a weighted sum of four domain-specific factors:
\begin{equation}\label{eq:relevance}
\text{RelScore}(q, c) = \sum_{i \in \{kw, type, loc, year\}} w_i \cdot S_i(q, c),
\end{equation}
where $S_{kw}$ denotes the IDF-weighted Jaccard similarity of keywords, $S_{type}$ and $S_{loc}$ represent binary matches for accident type and tribunal jurisdiction, and $S_{year} = 1/(1+|\Delta_{\text{year}}|)$ captures temporal proximity. We empiricallly set the weights $(w_{kw}, w_{type}, w_{loc}, w_{year})$ to $(0.5, 0.4, 0.05, 0.05)$ and define the \textit{pseudo-gold set} as candidates exceeding a $\text{RelScore}$ of 0.5.

Retrieval performance is assessed using \textbf{NormRecall@100} and \textbf{nDCG@100}. To account for varying sizes of the gold set across different accident patterns, NormRecall@$k$ measures coverage relative to the maximum possible recall achievable at a given cutoff $k$:
\begin{equation}\label{eq:normrec}
\text{NormRecall}@k = \frac{|Retrieved_k \cap GoldSet|}{\min(k, |GoldSet|)}.
\end{equation}
Complementarily, nDCG@100 evaluates the ranking quality by rewarding the placement of highly relevant candidates (those with higher $\text{RelScore}$) at earlier ranks. The DCG is computed as $\sum_{i=1}^{k} \frac{G_i}{\log_2(i+1)}$, where the gain $G_i$ is defined by the $\text{RelScore}$ of the candidate at rank $i$, subsequently normalized by the Ideal DCG (IDCG).

Regarding the generation task, the expert-authored \textit{Causes} field serves as the ground-truth reference for evaluating the synthesized cause narrative $n$. We employ a multi-faceted evaluation involving ROUGE-L for lexical overlap, SBERT cosine similarity for semantic alignment, and an LLM-as-a-judge score on a 1–5 scale to assess overall coherence and grounding \cite{zheng2023judging}. For the structured tags $T$, we report exact-match accuracy for the primary L1 group along with Micro-F1 and Jaccard scores for the multi-label L2 factors.

All retrieval and generation processes are executed on the original Korean adjudication text to preserve domain-specific linguistic nuances and technical precision. For the purpose of international dissemination, all tables and case examples provided in this paper utilize condensed English translations.

\section{RESULTS}
\subsection{Retrieval Results}
Table~\ref{tab:retrieval} shows that leveraging document fields is critical.
Whole-document indexing mixes accident description, causal reasoning, and administrative outcomes into one long text block, which dilutes field-specific signals and increases spurious matches.
Field-wise indexing enables the retriever to match scenario descriptions in Summary while also capturing technical causal patterns in \textit{Causes} and decision patterns in \textit{Disposition}, improving both coverage and ranking quality.

\textbf{Impact of field separation.}
Moving from whole-document indexing (Dense: 0.1819 NormRecall) to multi-field indexing over all fields (Dense: 0.4981) yields a 174\% relative gain in coverage.
Notably, adding \textit{Disposition} to \textit{Summary}+\textit{Causes} provides a large additional boost (NormRecall 0.3060 $\rightarrow$ 0.4981; nDCG 0.5336 $\rightarrow$ 0.7512), suggesting that administrative outcomes carry discriminative signals that help disambiguate lexically similar but causally different cases.

\textbf{Effectiveness of hybrid retrieval.}
Hybrid retrieval yields an additional gain by combining complementary strengths.
BM25 captures exact matches of rare legal/technical tokens, while dense retrieval captures paraphrases and semantically equivalent expressions.
Fusing both reduces misses caused by either lexical mismatch or over-reliance on surface keywords, improving NormRecall from 0.4981 (Dense) and 0.4215 (BM25) to 0.5463 (Hybrid).

\begin{table}[t]
\centering
\caption{Retrieval performance (top-$k$=100).}
\label{tab:retrieval}
\vspace{-0.1 in}
\footnotesize
\renewcommand{\arraystretch}{1.10}
\begin{tabularx}{\columnwidth}{@{} X c c @{} }
\toprule
\textbf{Strategy} & \textbf{Norm. Recall} & \textbf{nDCG} \\
\midrule
Whole-Document (Dense) & 0.1819 & 0.6123 \\
Single-Field Summary (Dense) & 0.1785 & 0.6055 \\
Single-Field Causes (Dense) & 0.1450 & 0.4405 \\
Multi-Field (Summary+Causes; Dense) & 0.3060 & 0.5336 \\
Multi-Field (All Fields; Dense) & 0.4981 & 0.7512 \\
Multi-Field (All Fields; BM25) & 0.4215 & 0.7034 \\
\textbf{Multi-Field (All Fields; Hybrid)} & \textbf{0.5463} & \textbf{0.7697} \\
\bottomrule
\end{tabularx}
\vspace{-0.1 in}
\end{table}

\subsection{Generation Results}
Table~\ref{tab:generation} reports end-to-end improvements when the generator is grounded in retrieved precedents (top-$k=40$ chunks).
The judge score increases substantially (3.340 $\rightarrow$ 3.723; +11.5\%), while SBERT similarity changes only marginally.
This pattern is expected: an instruction-tuned LLM can produce fluent summaries from the scenario alone, but retrieval primarily improves factual consistency and key-cause coverage by injecting concrete precedent evidence.

For tags, improvements are smaller but consistent (e.g., L2 micro-F1 0.460 $\rightarrow$ 0.490), indicating that retrieval provides useful signals for taxonomy alignment even without supervised tag training.
In practice, the largest qualitative benefit is that outputs become \emph{precedent-grounded}: the generator can cite retrieved causes/dispositions as support, which is not possible for an LLM-only baseline.

While traditional supervised classifiers can perform isolated tag prediction, they are structurally incapable of generating the cohesive causal narratives or providing the traceable precedent evidence required for this task. Therefore, to evaluate the end-to-end framework, we compare the proposed method against an LLM-only baseline, explicitly isolating the impact of retrieval grounding on structured generation.

\begin{table}[t]
\centering
\caption{Generation performance (Top-$k=40$).}
\label{tab:generation}
\vspace{-0.1 in}
\footnotesize
\renewcommand{\arraystretch}{1.10}
\begin{tabularx}{\columnwidth}{@{} X c c c @{} }
\toprule
\textbf{Metric} & \textbf{LLM-only} & \textbf{Proposed} & \textbf{Improv.} \\
\midrule
ROUGE-L (F1) & 0.192 & 0.206 & +7.3\% \\
Semantic Sim. (SBERT) & 0.723 & 0.724 & +0.1\% \\
LLM-as-a-Judge (1--5) & 3.340 & 3.723 & +11.5\% \\
L1 Accuracy & 0.771 & 0.774 & +0.4\% \\
L2 Micro-F1 & 0.460 & 0.490 & +6.5\% \\
L2 Jaccard & 0.365 & 0.395 & +8.2\% \\
\bottomrule
\end{tabularx}
\vspace{-0.1 in}
\end{table}

\subsection{Case Study}
Table~\ref{tab:case_study} highlights two representative cases demonstrating how multi-field retrieval grounding mitigates hallucinated causal links. 
In the LLM-only setting, the model frequently relies on spurious correlations derived from tangential scenario details. 
For instance, in the stranding case, the LLM-only baseline infers vague structural defects from the ``sudden impact'' description, completely missing the geographical hazard. 
In the explosion case, the presence of an intoxicated crew triggers an erroneous logical leap to alcohol vapors as the ignition source. 

Conversely, the proposed framework leverages retrieved precedents—specifically utilizing the technical and legal context preserved in the independent Causes and Disposition fields—to ground the generation process. 
This contextual injection shifts the model's reasoning from speculative deduction to precedent-aligned diagnostic inference. 
By accessing historically and situationally similar records, the framework correctly identifies specific technical mechanisms, such as charting failures and propane gas ignition, thereby producing an analysis highly consistent with expert judgments.

\begin{table*}[t]
\centering
\caption{Case Study Examples: Comparison of Causal Reasoning (Condensed)}
\label{tab:case_study}
\vspace{-0.1 in}
\footnotesize
\renewcommand{\arraystretch}{1.10}
\begin{tabularx}{\textwidth}{@{} p{0.22\textwidth} p{0.18\textwidth} X X @{} }
\toprule
\multirow{2}{*}{\textit{Scenario (Excerpt)}} & \multirow{2}{*}{\textit{Expert Causes}} & \multicolumn{2}{c}{\textit{Generated Analysis}} \\
\cmidrule(l){3-4}
& & \textit{LLM-only Baseline} & \textit{Proposed Framework} \\
\midrule
\textsc{[Stranding]} \newline Occurred shortly after passing a lighthouse in calm weather and good visibility; a sudden severe hull impact and shaking were reported. &
Approached a reef too closely without position verification; improper navigation near the hazard. &
Impact was likely due to structural defects in the hull or improper equipment handling; stability may have been weakened despite the clear weather. &
Vessel stranded on underwater rocks after navigating too close to the lighthouse; the core cause was failure to verify position and recognize reefs on the chart. \\
\midrule
\textsc{[Fire/Explosion]} \newline Occurred shortly after a crew member entered a cabin with matches and cigarettes; a sudden explosion and fire were reported, causing casualties. &
Negligent handling of propane gas; gas leaked and accumulated in the cabin and was ignited when a match was struck. &
Explosion was caused by alcohol vapors accumulated from drinking in the enclosed cabin; poor supervision of intoxicated crew was the primary factor. &
Explosion was caused by ignition of leaked propane gas accumulated in the cabin due to poor maintenance; the match acted as the ignition source. \\
\bottomrule
\multicolumn{4}{@{}p{\linewidth}@{}}{
  \scriptsize 
  \textit{* All content examples are translated from Korean.}
}
\end{tabularx}
\vspace{-0.2 in}
\end{table*}

\section{DISCUSSION AND LIMITATIONS}

The experimental results validate that structuring RAG architectures around domain-specific document fields significantly enhances downstream generation. By isolating descriptive narratives (\textit{Summary}) from technical and legal reasoning (\textit{Causes} and \textit{Disposition}), our multi-field hybrid retrieval strategy effectively reconstructs the diagnostic reasoning pathway utilized by expert investigators. This structural alignment ensures that the generator is grounded not merely in lexically similar texts, but in historically and causally analogous precedents. Furthermore, the fusion of dense and sparse retrieval signals proves essential for maritime safety workflows, providing robustness against vocabulary mismatches while preserving the precise retrieval of critical regulatory and technical terminology.

Despite these gains, certain limitations remain within the evaluation framework. Due to the absence of large-scale expert relevance labels, retrieval performance was measured using a metadata-based proxy score. Additionally, the hierarchical cause tags were assigned via rule-based extraction rather than manual annotation, and the LLM-as-a-judge evaluation carries the inherent risk of model-dependent bias. However, formulating this fixed, heuristic proxy directly addresses the practical lack of a universal gold standard in specialized administrative domains. It provides a reproducible, systematic baseline for large-scale strategy comparisons under realistic data constraints. Ultimately, the consistent end-to-end generation improvements confirm that optimizing for this retrieval proxy translates into more accurate and traceable RCA outputs, while highlighting future directions for enhancing the uncertainty-awareness of the evaluation framework \cite{yoon2024uncertainty}.

\section{CONCLUSION}
This paper proposed a multi-field hybrid RAG framework for maritime accident RCA, utilizing a comprehensive dataset of KMST adjudication reports. By structuring historical records into field-aware ``incident cards'' and fusing sparse and dense retrieval signals via RRF, the proposed method significantly enhances the quality of precedent retrieval. To address the lack of human-annotated gold standards in specialized domains, we introduced a reproducible, metadata-based evaluation proxy that enables a systematic and objective comparison of retrieval strategies. End-to-end experiments demonstrate that these retrieval enhancements directly translate into more accurate, structured, and precedent-grounded RCA generation compared to an LLM-only baseline. Our framework establishes a practical foundation for automated maritime investigation support. Future work will explore the integration of formal legal regulations, such as COLREGs, and graph-based reasoning to further expand the system's diagnostic depth and interpretability.

\section*{Acknowledgment}
This work was supported by the National Research Foundation of Korea(NRF) grant funded by the Korea government(MSIT)(No.RS-2023-00218913).
\bibliographystyle{IEEEtran}
\bibliography{references}
\end{document}